# Physics Enhanced Residual Policy Learning (PERPL) for safety cruising in mixed traffic platooning under actuator and communication delay

Keke Long, Haotian Shi, Yang Zhou, Xiaopeng Li*

Corresponding author email: xli2485@wisc.edu

**ABSTRACT**

Linear control models have gained extensive application in vehicle control due to their simplicity, ease of use, and support for stability analysis. However, these models lack adaptability to the changing environment and multi-objective settings. Reinforcement learning (RL) models, on the other hand, offer adaptability but suffer from a lack of interpretability and generalization capabilities. This paper aims to develop a family of RL-based controllers enhanced by physics-informed policies, leveraging the advantages of both physics-based models (data-efficient and interpretable) and RL methods (flexible to multiple objectives and fast computing). We propose the Physics-Enhanced Residual Policy Learning (PERPL) framework, where the physics component provides model interpretability and stability. The learning-based Residual Policy adjusts the physics-based policy to adapt to the changing environment, thereby refining the decisions of the physics model. We apply our proposed model to decentralized control to mixed traffic platoon of Connected and Automated Vehicles (CAVs) and Human-driven Vehicles (HVs) using a constant time gap (CTG) strategy for cruising and incorporating actuator and communication delays. Experimental results demonstrate that our method achieves smaller headway errors and better oscillation dampening than linear models and RL alone in scenarios with artificially extreme conditions and real preceding vehicle trajectories. At the macroscopic level, overall traffic oscillations are also reduced as the penetration rate of CAVs employing the PERPL scheme increases.

# 1 INTRODUCTION

With the development of Connected and Autonomous Vehicles (CAVs), there arises the necessity for effective controllers to navigate CAVs in mixed traffic scenarios of CAVs and Human-driven Vehicles (HVs) (Ghiasi et al., 2017; Yao and Li, 2020). Such mixed traffic scenarios are inherently complex and dynamic, primarily due to the unpredictable behaviors of HVs and other road users (K. Yang et al., 2023). This complexity presents a significant challenge for CAV controllers. Inadequate control strategies for CAVs in these environments can result in increased traffic congestion and heightened safety risks.

Mainstream methodologies for controlling CAVs can be classified into three primary types: closed-form controllers, controllers based on Model Predictive Control (MPC), and those utilizing Reinforcement Learning (RL). The closed-form linear (Li, 2022) state-feedback CAV controller has been favored due to its rapid and straightforward implementation facilitated by its analytical representation. However, despite their simplicity, these controllers struggle with constrained optimization frameworks involving multiple explicit objectives and constraints, making them less adaptable to the dynamic driving environments found in mixed traffic. The MPC-based controller, supported by a flexible framework, addresses this limitation by optimizing multiple objectives with constraints in a rolling horizon (Zhou et al., 2019). Nonetheless, MPC often necessitates convex problem formulations and imposes a relatively high computational burden, hindering high-resolution real-time implementation. Some learning-based MPCs employ neural networks to solve problems, which mitigates slow-solving issues. However, these systems still face inherent challenges within the MPC framework, as their optimizer is trained through imitation learning based on datasets of MPC solutions. This training approach inherently limits their performance to not exceed that of the quality of the underlying MPC controller (Sacks et al., 2023).

In contrast to MPC, RL-based control facilitates fast computing for real-time implementation (He et al., 2024; Qu et al., 2020). Furthermore, RL-based controllers can capture the nonlinear and stochastic characteristics of complex systems due to the capabilities of neural networks. However, as summarized from previous literature, RL encounters several challenges regarding applicability in mixed traffic scenarios. In environments where autonomous and human-driven vehicles interact, the unpredictability of human behavior adds complexity to the driving context. RL's inherent issues with generalization mean it might not reliably interpret or react to the diverse behaviors encountered among different drivers, which could lead to safety risks or inefficient traffic flow. Moreover, RL's challenges with safe exploration become particularly critical in mixed traffic, as the system should navigate safely without extensive prior exposure to every possible driving scenario, potentially leading to unsafe actions in unanticipated situations. These limitations underscore the need for robust testing and the integration of safety-oriented strategies within RL frameworks before deployment in dynamic, real-world environments.

Safe exploration stands out as a significant challenge for RL applications in the domain of autonomous vehicle control. The lack of safety assurance poses a major obstacle to RL application in real-world scenarios, manifesting in two key aspects. First, the generalization ability of RL has long remained an unsolved issue. Generalization requires RL models to be robust to variations in their environments and able to transfer and adapt to unseen (but similar) environments during deployment. However, current RL methods often evaluate the policy in the same environment it was trained in (Kirk et al., 2023). Over-reliance on data-driven approaches exposes the system to data biases, leading to overfitting of the training data. When deployed in the real world, RL and other deep learning models often encounter previously unseen categories of samples—out-of-distribution (OOD) data—posing significant challenges. This limitation hinders existing autonomous systems from effectively addressing long-tail and cross-domain issues, restricting their safety and adaptability in new environments (Li et al., 2023). Second, RL may learn behaviors that violate physical laws because it relies on function approximation and representation learning (Cao et al., 2024), making it heavily dependent on data quality. This is a common issue for learning-based models (Long et al., 2024). Thus, RL may not learn skills not provided or rare in the demonstration data. Moreover, many RL-based vehicle control approaches, such as Adaptive Cruise Control (ACC) and Cooperative Adaptive Cruise Control (CACC), address safety by incorporating it into the reward function, aka Lagrangian relaxation (Shi et al., 2023; Tang et al., 2024; Chen et al., 2022), converting hard constraints into penalty terms within the objective function. Consequently, until the policy converges, we cannot expect constraints to be fully considered and met (Chen et al., 2022; Y. Yang et al., 2023; Zhao et al., 2023). Ensuring vehicles explore new behaviors under strict safety constraints necessitates strict safety measures from the outset of training, which is challenging to implement and requires prior knowledge about the environment or task (Yue et al., 2024).

To address existing safety concerns in RL applications for autonomous vehicle control, incorporating model-based strategies provides the possibility to provide interpretable safe improvement. This research proposes PERPL: a Physics-Enhanced Reinforcement Learning framework with enhanced safety assurance. PERPL introduces two methods to incorporate physics prior information into RL. Firstly, from a policy perspective, PERPL adopts a Residual Action Policy, integrating the data-driven RL method with model-based decision-making, resulting in a residual action policy framework. This framework allows the model-based action policy to guide DRL agents' exploration during training while the DRL policy learns to effectively handle the uncertainties. Secondly, at the action level, this study innovatively introduces a Physics-Informed Safety Action Barrier, which works with the Residual Action Policy to enhance the safety measures within the PERPL framework. This Safety Action Barrier utilizes physics-based constraints to refine and adjust the actions proposed by the PERPL controller, ensuring that every action adheres to predefined safety parameters. By doing so, it provides a robust safety guarantee to unseen domains in mixed traffic scenarios. The efficacy of this framework is validated at both the individual vehicle and platoon levels, considering communication and actuator delays. The primary contributions of this paper are twofold:

1. The introduction of the PERPL framework in mixed traffic platooning, which synergizes a physics-based control policy with reinforcement learning.
2. A demonstration of how the PERPL framework's generalization capabilities on out-of-distribution data surpass those of standalone physics-based or RL approaches.

This paper is organized as follows. Section 2 outlines the investigated problem. Section 3 introduces the proposed PERPL controller. Section 4 presents the experiments conducted to compare the proposed model with baseline models. Finally, Section 5 provides the conclusion and discusses directions for future research.

## 2  PROBLEM STATEMENT

### 2.1  Environment Setting

This research focuses on the longitudinal control of CAVs in mixed traffic of CAVs and HVs. We consider the car following process without lateral movement on the highway. We denote $\mathcal{N}$ as the set of vehicle index, $\mathcal{N} = \{\mathcal{N}^{CAV}, \mathcal{N}^{HV}\}$, where $\mathcal{N}^{CAV}$ is the set of CAVs and $\mathcal{N}^{HV}$ is the set of HVs. CAVs broadcast their state information (e.g., speed, position) to other vehicles in the platoon via Vehicle-to-Vehicle (V2V) communications, subject to a constant communication delay. They can also access real-time data about their own state. HVs lack autonomous driving capability and do not receive digital information from other vehicles, but HVs could broadcast information to CAVs within communication range.

This research considers a communication delay $\tau^c$ during receiving preceding vehicle data, which includes both the signal propagation time and the processing time once the data is received. Moreover, when executing decisions derived from the PERPL framework, the vehicles experience actuator delay $\tau^A$, which refers to the time lag between the issuance of commands by the control system and the actual response by the vehicle's actuators. The related notations are defined in TABLE 1.

### 2.2  Distributed Platoon Control Scheme

In this section, we proposed a distributed CAV longitudinal control for distributed platooning, whose framework is presented in Figure 1. Any CAV, denoted as $n \in \mathcal{N}^{CAV}$, can obtain its own state at time $t$: $\boldsymbol{s}_{nt}, n \in \mathcal{N}^{CAV}, t \in \mathcal{T}$, where $\mathcal{T}$ denotes the set of all time stamps. each CAV can access the states of up to three preceding vehicles within the communication range. The states of these vehicles are represented as $\{s_{(n-k)(t-\tau^C)}\}_{k\in[1,2,3]}, n \in \mathcal{N}^{CAV}$, where $\tau^C$ is the communication time delay. If fewer than three vehicles precede any CAV or if there are no preceding vehicles at all, the number of vehicles considered adjusts accordingly to zero, one, or two based on availability. This approach ensures that each vehicle can dynamically adjust its behavior based on the immediate traffic conditions, enhancing both the responsiveness and safety of the platoon.

Vehicle state $s_{nt}$ including three parts: position error $\Delta d_{nt}$, speed difference $\Delta v_{nt}$ and acceleration $a_{nt}$: $s_{nt} = [\Delta d_{nt}, \Delta v_{nt}, a_{nt}]^T, n \in \mathcal{N}, t \in \mathcal{T}$. Definitions for these components are as follows:

CAVs apply a constant time headway (CTH) spacing strategy, which follows its preceding vehicle with a desired spacing distance and ensures safety. Thus, the desired spacing distance of vehicle $n$ at time $t$ is $h^d v_{nt} + d_0$, where $h^d$ and $d_0$ are the desired constant headway and standstill space, respectively. Based on the CTH rule, the position error of Vehicle $n$ with respect to a desired distance from the preceding vehicle $(n-1)$ was denoted by $\Delta d_{nt}$:

$$\Delta d_{nt} = d_{(n-1)(t-\tau^C)} - d_{nt} - d_0 - h^d v_n, n \in \mathcal{N}, t \in \mathcal{T} \tag{1}$$

The speed difference between the ego and the preceding vehicle is:

$$\Delta v_{nt} = v_{(n-1)(t-\tau^C)} - v_{nt}, n \in \mathcal{N}, t \in \mathcal{T} \tag{2}$$

The control variable is $u_{nt}$. Given the assumptions and communication environment, the vehicle dynamics are modeled by linearized dynamics with the consideration of time delay $\tau^A$:

$$\dot{a}_{nt} = \frac{u_{nt} - a_{nt}}{\tau^A}, n \in \mathcal{N}, t \in \mathcal{T} \tag{3}$$

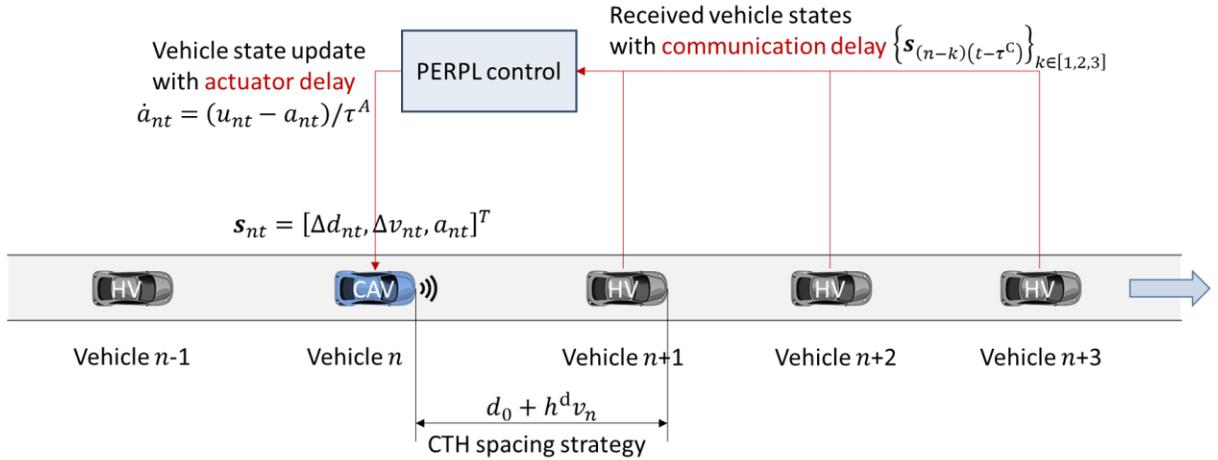

Figure 1 Distributed control scheme for vehicular platoon.

Table 1 Key notation

| Notation | Description |
|---|---|
| $\mathcal{N}$ | Set of vehicle trajectories |
| $n$ | Index of trajectory. $n \in \mathcal{N}$ |
| $\mathcal{T}$ | Set of all time stamps |
| $t$ | Index of time $t \in \mathcal{T}$ |
| $d_{nt}$ | Longitudinal position of Vehicle $n$ at time $t$ $(m)$ |
| $v_{nt}$ | speed of Vehicle $n$ at time $t$ $(m/s)$ |
| $a_{nt}$ | Acceleration of Vehicle $n$ at time $t$ $(m/s^2)$ |
| $h^d$ | Desired headway $(s)$ |
| $d_0$ | desired space at a standstill (m) |
| $\Delta d_{nt}$ | Distance error of vehicle $n$ under CTH spacing strategy at time $t$ $(m)$ |
| $\Delta v_{nt}$ | speed difference of Vehicle $n$ at time $t$ (m/s) $\Delta v \coloneqq v_{n-1} - v_n$ |
| $\boldsymbol{s}_{nt}$ | state of Vehicle $n$ at time $t$, $\boldsymbol{s}_{nt} \coloneqq [\Delta d_{nt}, \Delta v_{nt}, a_{nt}]^T$ |
| $\tau^A$ | Actuator time delay $(s)$ |
| $\tau^C$ | Communication time delay $(s)$ |
| $\mathcal{A}$ | Feasible set of $a_{nt}$. |
| $\boldsymbol{a}_{nt}^{\text{phy}}$ | Action given by physics model-based action policy |
| $\boldsymbol{a}_{nt}^{\text{RL}}$ | Action given by RL-based residual action policy |
| $f^{\text{SAG}}$ | Physics-based safety action barrier |
| $d_n$ | Damping ratio of vehicle $n$ |

## 3 PERPL CONTROLLER

### 3.1 Framework

Our proposed control framework encompasses two parallel controllers: a linear controller that focuses on ensuring local and string stability using a non-linear programming formulation and a DRL controller that specifically targets handling traffic disturbances and time delays. By integrating these two controllers, our framework aims to effectively address the challenges associated with car-following control in CAV environments. A safety barrier is added to guarantee safety. As shown in Figure 2, the control action $\boldsymbol{a}_{nt}^{\text{PERPL}}$ from PERPL framework is:

$$\boldsymbol{a}_{nt}^{\text{PERPL}} = f^{\text{SG}}\left(\boldsymbol{a}_{nt}^{\text{phy}} + \boldsymbol{a}_{nt}^{\text{RL}}\right) \quad (4)$$

where $\boldsymbol{a}_{nt}^{\text{phy}}$ is the output action of physics-based policy, $\boldsymbol{a}_{nt}^{\text{RL}}$ is the output action of residual policy. $f^{\text{SG}}$ is the safety barrier that projects the combined output $\boldsymbol{a}_{nt}^{\text{phy}} + \boldsymbol{a}_{nt}^{\text{RL}}$ to a safety range to guarantee safety (Ames et al., 2017). The details of each component are given in the following subsections.

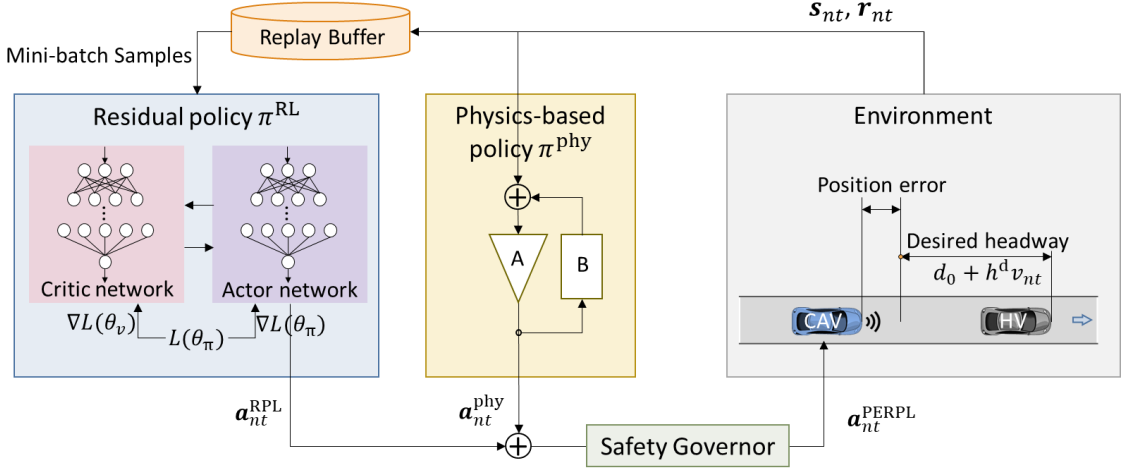

Figure 2 PERPL Framework.

## 3.2 Model-based Policy

A linear control model is employed to manage the behavior of a following vehicle within a platoon, emphasizing maintaining constant headway and minimizing the deviation from the lead vehicle's trajectory. The model is based on principles of classical control theory. The control input from the linear controller is:

$$\boldsymbol{a}_{nt}^{\text{phy}} = K \cdot \boldsymbol{s}_{nt} \tag{5}$$

where $K = [K_d, K_v, 0]$ is a vector of feedback gained forming a closed loop of the controller, where $K_d, K_v$ are the feedback coefficients for $\Delta d_{nt}$ and $\Delta v_{nt}$.

## 3.3 Residual Policy

The residual action policy is constructed using the Proximal Policy Optimization (PPO) method (Schulman et al., 2017). Unlike other reinforcement learning methods that use the sum of discounted rewards to estimate future returns, the PPO method employs a policy gradient method that focuses on optimizing an objective function. This function is not solely based on the sum of returns but also involves the concept of probability ratios of policies and advantage functions, which facilitates learning and exploration on top of the existing policy. Therefore, PPO attempts to maintain training stability by limiting the extent of policy updates, thereby preventing excessively large updates during training.

The PPO is based on the Actor-Critic structure, with an actor (policy) and critic (value) network. This dual network structure facilitates efficient exploration of the action space while stabilizing the learning updates through the critic's value estimates.

### 3.3.1 Actor network

The Actor network is responsible for defining the policy $\pi$ with parameter $\theta$. It takes the DRL state as input and outputs a probability distribution over actions. The control signal $\boldsymbol{a}$ is then sampled from this distribution. The Actor network is updated by maximizing the objective function $L(\boldsymbol{s}, \boldsymbol{a}, \theta_t, \theta)$, which is defined as:

$$L(\boldsymbol{s}, \boldsymbol{a}, \theta_t, \theta) = \min\bigl(r_t(\theta)\hat{A}_t, \text{clip}(r_t(\theta), 1-\epsilon, 1+\epsilon)A_t\bigr) \tag{6}$$

where $r_t(\theta)$ is the probability ratio $r_t(\theta) = \pi_\theta(a|s)/\pi_{\theta_t}(a|s)$, $\text{clip}(\cdot)$ is a clipping function to remove incentives for the new policy to get far from the old policy, which prevents large updates that could destabilize the training process. $\epsilon$ is a small hyperparameter, which determines how much the ratio can differ from 1 before it is clipped. $\hat{A}_t$ is the advantage estimate at time $t$, calculated by:

$$\hat{A}_t = R_t - V(s_t) \tag{7}$$

where $V(s_t)$ is the value estimated by the Critic network; $R_t$ denotes the discounted sum of rewards over T steps at state $s_t$:

$$R_t = r_t + \gamma r_{t+1} + \cdots + \gamma^{T-t+1} r_{T-1} + V^{T-t}(s_T) \tag{8}$$

Therefore, the parameter $\theta$ of the Actor network is updated based on the gradient of $L(s, a, \theta_t, \theta)$ with a learning rate $\alpha^\theta$. This update method effectively balances policy improvement with training stability, using a clipping mechanism to prevent excessively aggressive policy updates in certain situations, which could otherwise lead to performance deterioration.

$$\theta = \theta - \alpha^\theta \cdot \nabla L(s, a, \theta_t, \theta) \tag{9}$$

$$\theta_{t+1} = \arg\max_\theta \mathbb{E}_{s,a \sim \pi_{\theta_t}}[L(s, a, \theta_t, \theta)] \tag{10}$$

### 3.3.2 Critic network

The Critic network evaluates the decision output by the Actor network. The Critic network receives the DRL state $s$ as input and outputs the estimated state value $V(s_t)$. The network structure also includes one hidden layer with 100 neurons, and the ReLU function is used as the activation function for the output. The Critic network is updated by minimizing the critic loss function:

$$L_c(\Phi) = \hat{E}_t[(V(s_t) - R_t)^2]$$

The parameter $\Phi$ is iteratively optimized based on the gradient $L_c(\Phi)$ with learning rate $\alpha^\Phi$:

$$\Phi = \Phi - \alpha^\Phi \cdot \nabla L_c(\Phi)$$

The residual action policy pseudocode is shown in TABLE 2.

TABLE 2 Pseudocode

| **Algorithm 1**: Training process of Residual action policy algorithm for Longitudinal cruising control |
|---|
| **Inputs:** state information, action information |
| Initialize agent network with critic and actor networks, environment, actor policy µ: S → R^m+1 and σ: S → diag(σ1, σ2, ..., σm+1) |
| **for** iteration=1,2,… **do** |
|     **for** actor=1,2,…N **do** |
|         Run policy $\pi_{\theta_t}$ in the environment for $T$ timesteps and collect $(s_t, a_t, r_t)$ |
|         Compute advantages estimates $\hat{A}_1, \ldots, \hat{A}_T$ with discount along the time |
|     **end for** |
|     Optimize surrogate objective $L$ with respect to $\theta$ by Eq. (6), with $K$ epochs and minibatch size $M \leq NT$ |
|     Update critic $\theta_t \leftarrow \theta_{t+1}$ |
| **end for** |

### 3.4 Physics-based Safety Barrier

The safety barrier projects the action into the safety region based on the safety requirement on headway. The safety barrier adjusts the control action by taking the combined output of the physics-based and RL policies, $a_{nt}^{\text{phy}} + a_{nt}^{\text{RPL}}$, which may not inherently satisfy safety requirements. It then computes the adjusted action $a_{nt}^{\text{PERPL}}$ that minimizes the deviation from this combined action while ensuring it adheres to safety constraints, formalized as:

$$a_{nt}^{\text{PERPL}} = f^{\text{SG}}\left(a_{nt}^{\text{phy}} + a_{nt}^{\text{RPL}}\right) = \arg \min_{a_{nt} \in \mathcal{A}} \left\| a_{nt} - \left(a_{nt}^{\text{phy}} + a_{nt}^{\text{RPL}}\right) \right\|^2 \tag{11}$$

subject to the constraint ensures that the resulting state transition remains within a safe state $s_{n(t+1)}^{\text{safe}}$:

$$A s_{nt} + B a_{nt} + C \mathbf{w}_{nt} \in s_{n(t+1)}^{\text{safe}} \tag{12}$$

where $s_{n(t+1)}^{\text{safe}}$ is the set of safety states where the headway between Vehicle $n$ and Vehicle $(n-1)$ is kept within safe limits, typically between 1s and 3s. The lower limit is to guarantee safety (Vogel, 2003), and the upper limit is to avoid overlarge headway causing unstable inside the platoon. $\mathcal{A}$ is the feasible set of $a_{nt}$.

Importantly, the safety barrier is not activated under normal safe control scenarios and does not compromise the control effectiveness of the model. While the safety barrier is not inherently a part of the PERPL framework, it is included for two main reasons: firstly, safety barriers are a common mechanism currently implemented in RL-based models in practical applications. Secondly, it helps account for physical constraints in real-world scenarios. In reality, a collision will occur if the following vehicle is too close to the leading vehicle. However, in simulation environments, the following vehicle may even pass the leading vehicle without any constraint. The safety barrier has been implemented to prevent such unrealistic scenarios. Although actual accidents might be mitigated, the frequency of activation of the safety barrier still serves as an important safety metric.

## 4 EXPERIMENTS

### 4.1 Experiment settings

This section details the design philosophy of the experiment, evaluation metrics, simulation methods for HVs in mixed traffic, and baseline models.

#### 4.1.1 Experiment Overview

Our analysis spans two dimensions: 1) **Single Vehicle Cruising**: We conducted training and testing using both artificially designed and real-world leading vehicle trajectories. The artificially designed trajectories were designed to create extreme danger scenarios (rapid acceleration and deceleration) to evaluate the model's generalization capabilities. Real-world trajectories were used to align more closely with practical conditions. 2) **Mixed Traffic Platooning**: We assessed the performance of multiple vehicles and conducted a macro-level analysis.

Communication delay $\tau_n^C$ was set at 0.3 seconds, based on empirical results, which lend a reasonable degree of reliability (Liang et al., 2024). The safety headway interval is defined between 1 and 3 seconds. The lower limit of headway prevents collisions, while the upper limit ensures vehicles do not lag too far behind the lead vehicle, thereby affecting subsequent vehicles. The safety barrier utilizes this range; if a pending action is detected that would cause the headway to exceed these bounds, the safety barrier projects the action to bring the headway back within a safe range. The desired constant headway $t^d$ and the safety headway set can be adjusted based on practical applications. Other parameters are listed in TABLE 3.

TABLE 3 Experiment parameters.

|  | **Parameters** | **Value** |
| --- | --- | --- |
| Experiment parameters | Vehicle length | 4 m |
|  | Update interval $\Delta t$ | 0.1s |
|  | Desired constant headway $h^d$ | 2s |
|  | The range of the safety headway | [1,3] s |
|  | Desired standstill space $d_0$ | 4 m |
|  | Actuator delay $\tau^A$ | 0.2 s |
| HV modeling parameters | Communication delay $\tau^C$ | 0.3 s |
|   | Desired velocity $V_0$ | 20.3 m/s |
|   | Safe time headway $T$ | 1.2 s |
|   | Maximum acceleration $a$ | 1.9 m/s² |
|   | Comfortable Deceleration $b$ | 3.9 m/s² |
|   | Acceleration exponent $\sigma$ | 4 |
|   | Minimum distance $S_0$ | 2.0 m |

*4.1.2 Evaluation metrics*

To systematically evaluate the proposed framework, three performance indicators to quantitatively assess the control performance: driving comfort.

**Headway Error:** To assess the stability and safety of vehicle operation, it is crucial that the headway—the distance between vehicles—remains consistent at the designated desired headway. The headway error reflects the vehicle's adherence to the Constant Time Gap (CTG) rule as well as its safety characteristics. For this purpose, we employ the headway's Root Mean Square Error (RMSE) to quantify deviations from the desired values.

$$\text{RMSE}^h = \sqrt{\sum_{t \in \mathcal{T}} \left( \frac{d_{(n-1)t} - d_{nt} - d_0}{v_{nt}} - h^d \right)^2 / |\mathcal{T}|} \tag{13}$$

**Damping Ratio:** The cumulative damping ratio ($d_i$) is a measure of the CAV controller's ability to dampen traffic oscillations, which quantifies the empirical string stability (Ploeg et al., 2014). When the traffic passes through a string-stable CAV, the magnitude of traffic oscillations is either reduced or remains unchanged. The $l_2$-norm acceleration damping ratio $d_n$ can be formulated as follows:

$$d_n = \frac{\|a_n\|_2}{\|a_0\|_2} = \left( \frac{\sum_{t \in \mathcal{T}} |a_{nt}|^2}{\sum_{t \in \mathcal{T}} |a_{0t}|^2} \right)^{1/2} \tag{14}$$

**Comfort level:** the driving cost function $c_{nt}$ aims to evaluate the eco-driving performance and empirical string stability, which is defined as:

$$c_{nt} = \alpha_i (a_{nt})^2 \tag{15}$$

*4.1.3 HV Modeling Method*

To ensure the simulation reflects realistic mixed traffic conditions, this study employs the Intelligent Driver Model (IDM), which is calibrated to depict the string instability characteristic often observed in HVs. This approach allows for a more convincing representation of HV behaviors within our experiments with parameters listed in TABLE 3.

*4.1.4 Baseline models*

To analyze the performance and influence of the PERPL framework in a mixed traffic environment, the Linear control and RL models are employed as a baseline for comparison. The Linear control model, similar to a model-based action policy, is calibrated on the training set with parameters set at $K_d = 0.62$ and $K_v = 0.37$. The RL (PPO) model shares a similar structural setup with the residual action policy used in the PERPL framework. These comparisons highlight the distinct advantages of integrating physical principles with RL techniques in managing complex traffic dynamics. Other training settings are listed in TABLE 4.

TABLE 4 Hyper Parameters of the PPO training

| Parameter | Value |
|---|---|
| Initial learning rate | 2e-4 |
| Discount factor $\gamma$ | 0.9 |
| gae_$\lambda$ | 0.95 |
| Update epochs | 10 |
| Clip ratio $\epsilon$ | 0.2 |
| Learning rate for value function optimizer | 0.5 |
| max_grad_norm | 0.5 |

## 4.2 Results of Single vehicle cruising

In this section, we compared three different control approaches for a single vehicle following. This setup helped us evaluate the accuracy and generalization capabilities of the PERPL model, compared to two baseline models: RL (PPO), and PERPL (Linear+PPO).

### 4.2.1 Data

In the single vehicle cruising scenario, the data comprises a mix of real-world trajectories from NGSIM (NGSIM, 2007) and a subset of artificially designed trajectories. Trajectories are divided into three sets: training set, test set, and extrapolation set, with each set containing 100 trajectories, containing 500 timesteps with a $0.1s$ time interval. Each controller is trained on each trajectory for 3000 episodes.

**Training set and Test set:** These are derived from NGSIM trajectories, with acceleration magnitudes limited to within $\pm 3 \ m/s^2$. The datasets are randomly split into training and testing subsets.

**Extrapolation set:** This set includes trajectories from NGSIM, modified to incorporate extreme acceleration and deceleration events beyond $\pm 3 \ m/s^2$. This arrangement allows for a comparative assessment of the generalization capabilities of various methods beyond their training datasets. Notice that the notably darker lines at $a = -4m/s^2$ and $a = 3m/s^2$. These were intentionally set to simulate extreme scenarios by artificially enhancing the acceleration and deceleration behaviors based on real driving trajectories. This approach is used to test the model's stability under challenging conditions.

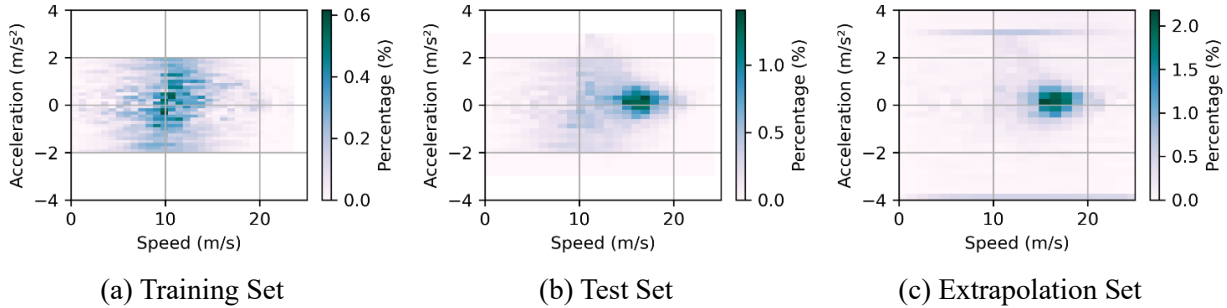

(a) Training Set    (b) Test Set    (c) Extrapolation Set

Figure 3 Distribution of acceleration and speed in three preceding vehicle trajectory sets.

### 4.2.2 Control Performance Evaluation

As shown in TABLE 5, both RL and PERPL achieved lower headway errors than the Linear controller in both training and test sets. In the extrapolation set, PERPL significantly outperformed the other methods, with headway errors much smaller than those of RL, which were six times higher than those of PERPL. An analysis of the Safety Barrier activation revealed that both PERPL and the Linear controller did not activate the Safety Barrier in any of the scenarios, demonstrating their robustness and consistent achievement of constant headway across all domains. However, the RL approach showed instability in the extrapolation set, with the Safety Barrier being activated 7.92% of the time, indicating that the RL model's control outputs occasionally pushed the state beyond safe limits. Figure 5 illustrates the vehicle following behavior under the three control models for a trajectory from the extrapolation set. Around the 130s mark, when the lead vehicle abruptly decelerated to -2.7, both Linear and PERPL managed to prompt rapid deceleration in the following vehicle, reaching about -3, whereas the RL controller only achieved a deceleration of around -1.7. This lesser response gradually increased headway error until the safety headway limit of 1s was breached around the 170s mark, triggering the Safety Barrier. This forced the vehicle to adopt an extremely sharp deceleration of -12.1 m/s² to exit the unsafe condition, preventing a potential collision and resulting in a greater average Headway RMSE.

TABLE 5 Single vehicle following performance artificially designed preceding trajectory.

|  | Average Headway RMSE $RMSE^h$ (s) | | | Proportion of time Safety barrier is activated (%) | | |
|---|---|---|---|---|---|---|
|  | Training set | Test set | Extrapolation set | Training set | Test set | Extrapolation set |
| Linear | 0.326 | 0.372 | 1.726 | 0 | 0 | 0 |
| RL (PPO) | 0.169 | 0.172 | 2.429 | 0 | 0 | 7.92 |
| PERPL (Linear+PPO) | 0.098 | 0.149 | 0.419 | 0 | 0 | 0 |

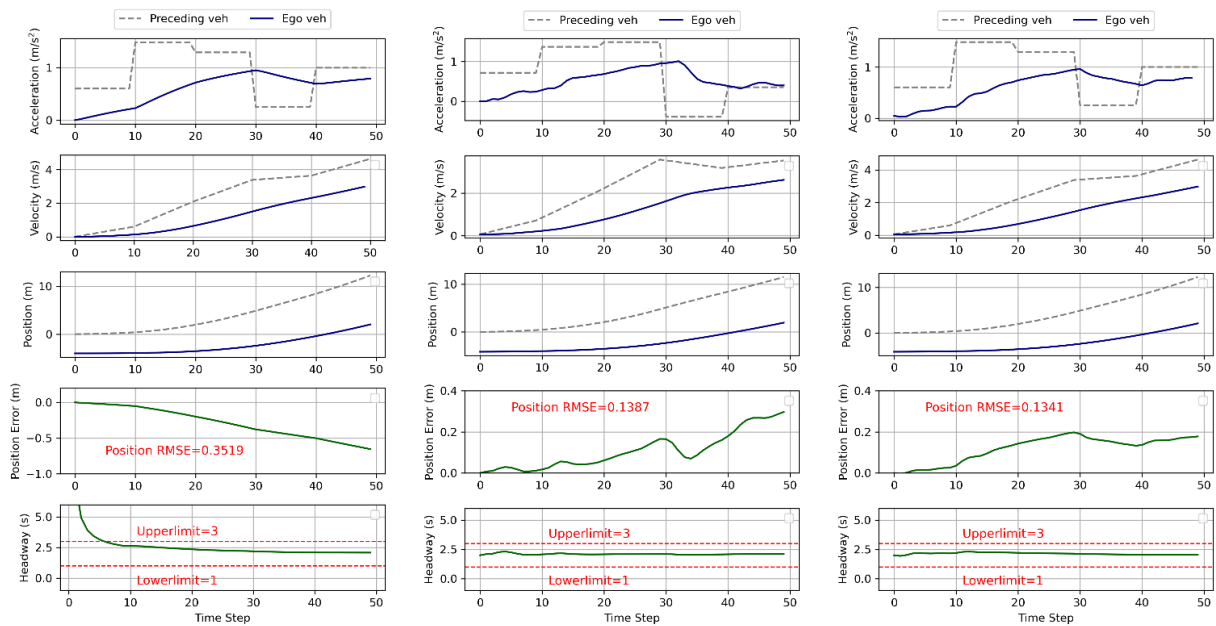

(a) Linear      (b) RL (PPO)      (c) PERPL (Linear+PPO)

Figure 4 Single vehicle following result of one example from the test set.

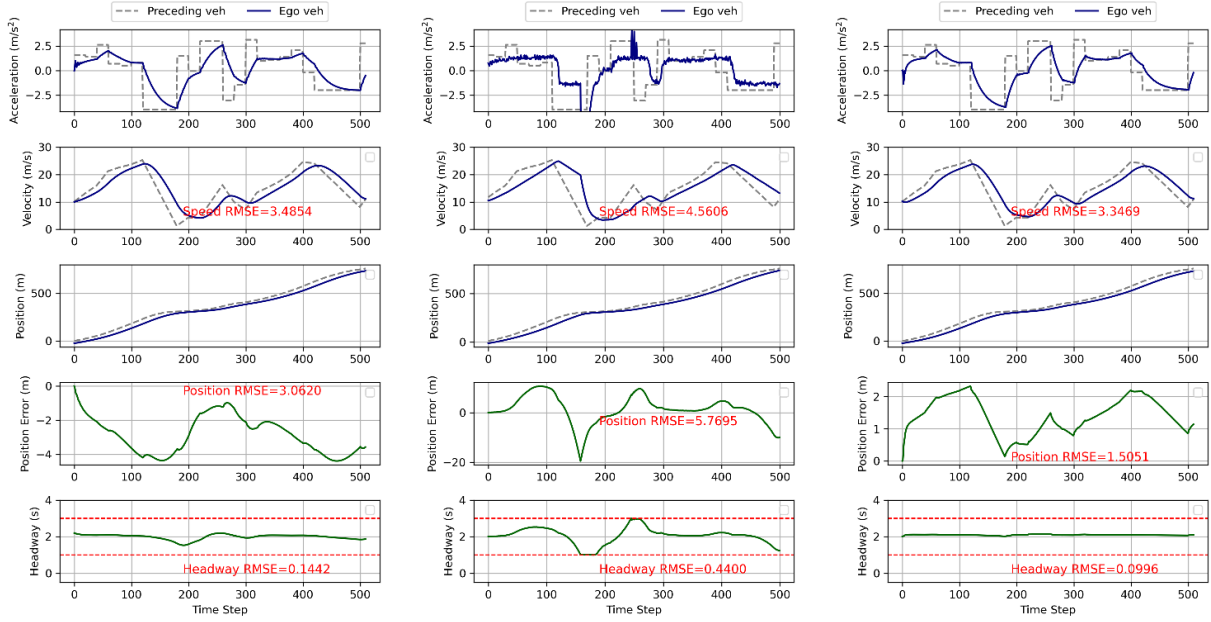

Figure 5 Single vehicle following result of one example from the extrapolation set.

To provide a clearer comparison of the behavior of the three actors in a wider range of scenarios, Figure 6 displays the action outcomes for each method under assumed conditions of zero acceleration with position error $\Delta d_{nt}$ ranging from -5 to 5 and speed difference $\Delta v_{nt}$ also ranging from -5 to 5. As depicted in Figure 7 (a), the Linear model's acceleration responses are proportional to changes in $\Delta d\_nt$ and $\Delta v\_nt$, indicating predictable behavior even under extreme conditions. Figure 7 (b) shows that the RL model's responses are not linear; it behaves similarly to the Linear model when $\Delta d_{nt}$ and $\Delta v_{nt}$ are small but adopts minimal absolute acceleration values under small $\Delta d_{nt}$ and $\Delta v_{nt}$, indicating a lack of aggressive response in critical situations. Conversely, the PERPL model in Figure 7 (c) exhibits behavior similar to the Linear model when $\Delta d_{nt}$ and $\Delta v_{nt}$ are within [-1,1] and adopts accelerations greater than 4 or decelerations less than -4 when the absolute values of $\Delta d_{nt}$ and $\Delta v_{nt}$ exceed 3, demonstrating its ability to respond assertively under extreme conditions.

Consider an extreme scenario where $\Delta d_{nt} = 5$ and $\Delta v_{nt} = 5$, indicating a situation where, if the lead vehicle's speed is $10 m/s$ and the following vehicle's speed is $15 m/s$, to maintain a 2s headway, the required distance would be $30\ m$. However, the actual distance is only $27\ m$, resulting in a headway of approximately 1.53 s and a Time-to-Collision (TTC) of 5.4 $s$, clearly necessitating rapid deceleration. Here, the RL model's deceleration rate is $-2.1\ m/s^2$, whereas the Linear model and PERPL achieve deceleration rates of approximately $4\ m/s^2$ and 5.3 m/s², respectively, highlighting the enhanced responsiveness of the PERPL model under critical conditions.

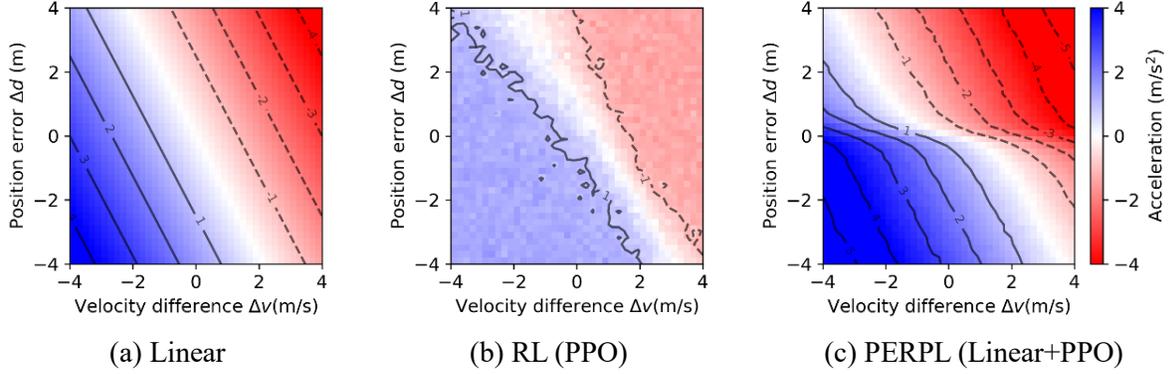

(a) Linear          (b) RL (PPO)          (c) PERPL (Linear+PPO)

Figure 6 Comparison of control policy trained on the training set.

To further clarify the differences in policy between RL and PERPL, Figure 7 shows how the control policies of these two models evolved during training. The training process of PPO is less stable than PERPL. In the first 3000 epochs, PPO incurred a higher loss (i.e., lower reward) compared to PERPL. From 3000 to 6000 epochs, after changing the leading vehicle data, PPO's performance significantly deteriorated relative to PERPL on the new data.

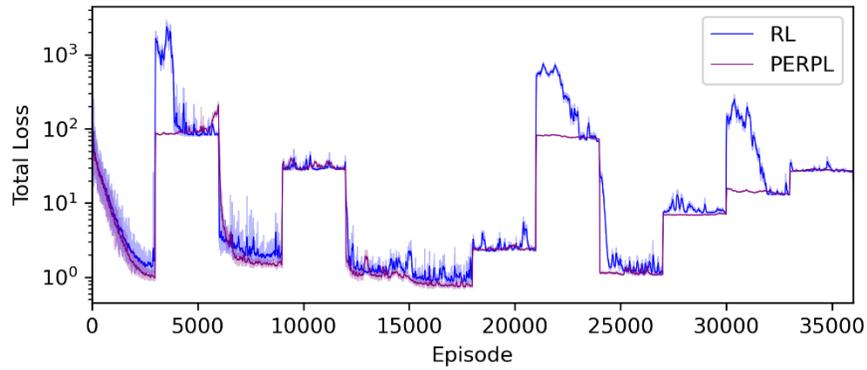

Figure 7 Total loss (Negative of rewards) during training.

Figure 8 further illustrates the control policies learned by PPO and PERPL across different epochs. PERPL controller starts with a linear pattern and gradually learns a non-linear policy. During training with various trajectories, only the results in the second quadrant ($\Delta v > 0 \text{ and } \Delta d < 0$) and the fourth quadrant ($\Delta v < 0 \text{ and } \Delta d > 0$) change slightly. Meanwhile, results in the first quadrant ($\Delta v > 0 \text{ and } \Delta d > 0$) and the third quadrant ($\Delta v < 0 \text{ and } \Delta d < 0$) remain largely consistent.

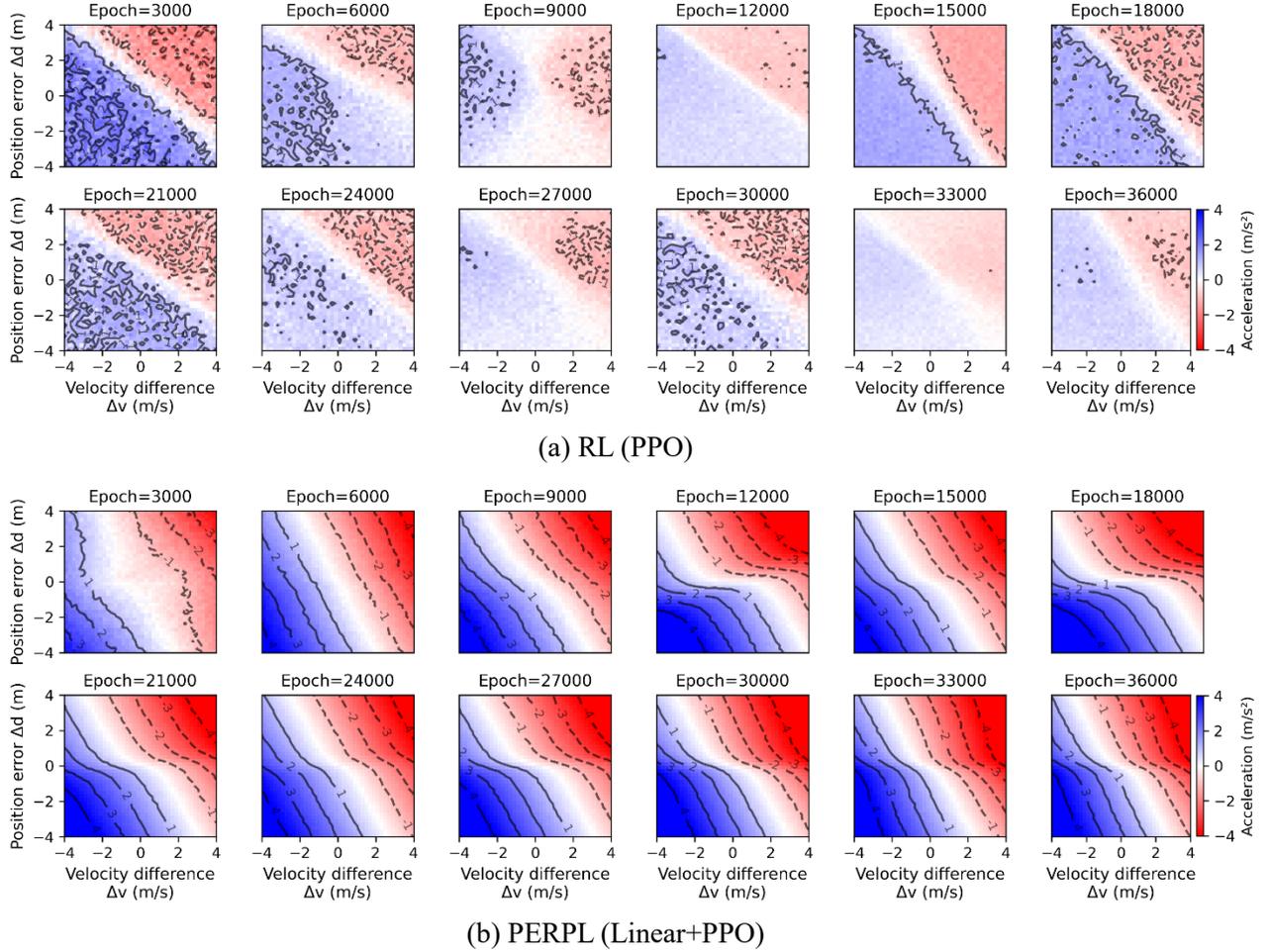

Figure 8. Control policy of RL and PERPL model during training process

### 4.3 Results of Mixed-traffic platooning

*4.3.1 Data*

Experiments utilizing real-world trajectory datasets are conducted to evaluate the Distributed Reinforcement Learning (DRL)-based control strategy. We constructed a platoon consisting of ten vehicles, labeled from upstream to downstream as 0, 2, 3, 4, 6, 7, and 9 as HVs and the remaining as CAVs. The trajectories for vehicle 0 on I-80 from 4:00 p.m. to 4:15 p.m., a period noted for frequent traffic oscillations, were selected for these experiments. To ensure consistency across trials, each vehicle in the experiment starts from an initial equilibrium state.

*4.3.2 Mixed platoon control performance*

Results illustrated in TABLE 6 and Figure 9 of the section reveal that the PERPL model outperformed the other models in terms of headway maintenance, with a lower RMSE, suggesting better precision in following distances. Additionally, the PERPL model showed improved damping ratios and comfort scores, indicating enhanced overall platoon stability and passenger comfort compared to the Linear and standalone RL models. These findings underscore the potential of integrating physics-based control with RL techniques to enhance automated driving systems in complex traffic environments.

As an illustrative example, Figure 10 shows the ten-vehicle platoon trajectories of the field data and simulated results using the proposed PERPL and baseline models. It can be seen that under linear control, the following vehicles are most affected by the lead vehicle's stop, with nearly all following vehicles coming to a stop after the lead vehicle stops.

TABLE 6 Mix-platoon performance.

|  | Headway RMSE ($s$) | Damping Ratio | Comfort Score |
|---|---|---|---|
| Linear | 0.439 | 0.616 | 0.301 |
| RL (PPO) | 0.233 | 0.575 | 0.263 |
| PERPL (Linear+PPO) | 0.204 | 0.558 | 0.249 |

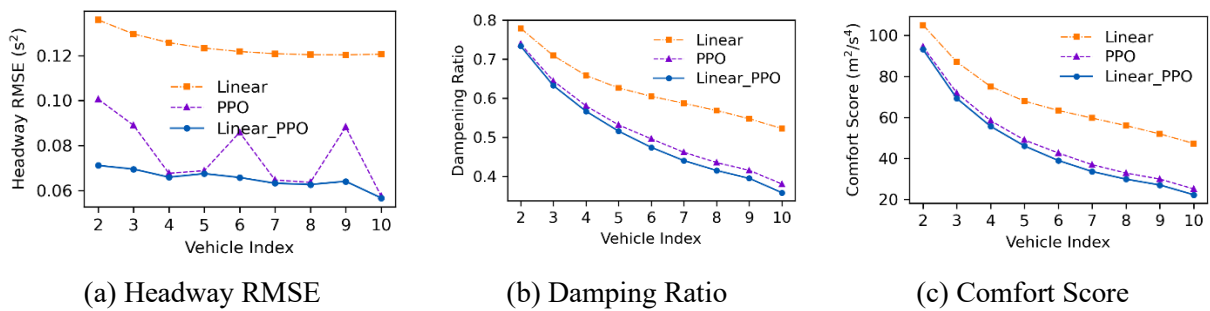

(a) Headway RMSE  (b) Damping Ratio  (c) Comfort Score

Figure 9 the metric results of each vehicle in the mixed platoon.

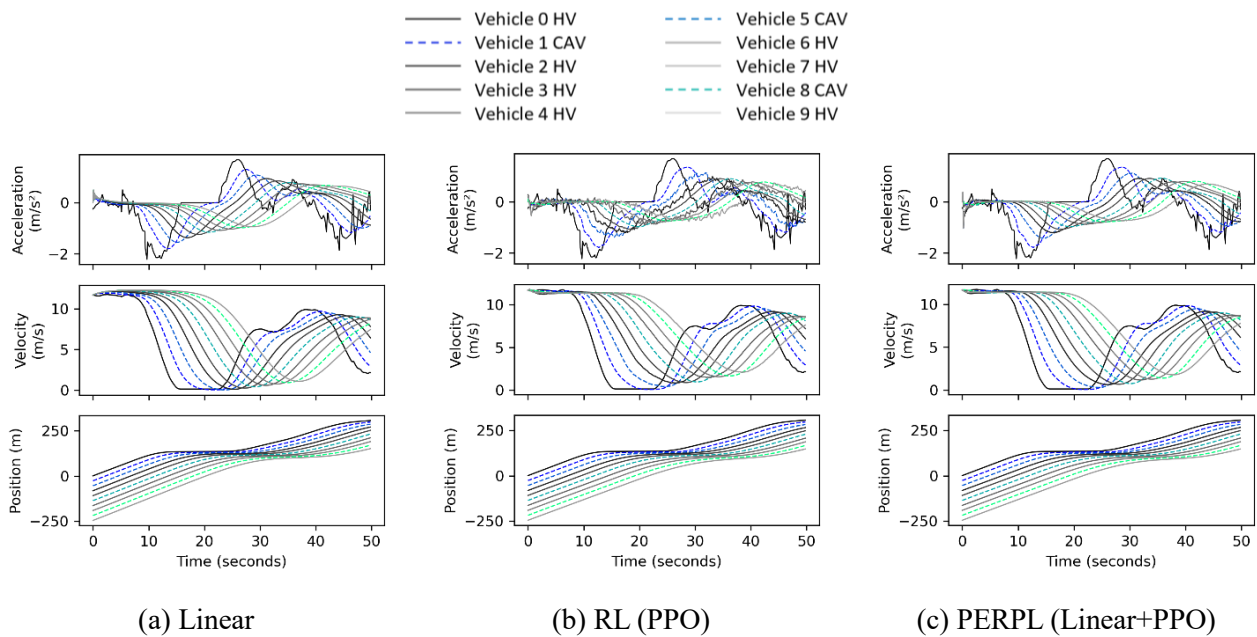

(a) Linear  (b) RL (PPO)  (c) PERPL (Linear+PPO)

Figure 10 The position, velocity, and realized acceleration results of mixed platoon.

### 4.3.3 Mixed platoons with different penetration rates

To visually demonstrate the dampening effectiveness of the proposed control strategy, we applied the proposed method in a 40-follower mixed platoon with different penetration rates (0%, 20%, 40%, 60%, 80%, 100%). In these scenarios, the CAVs were randomly distributed throughout the mixed traffic.

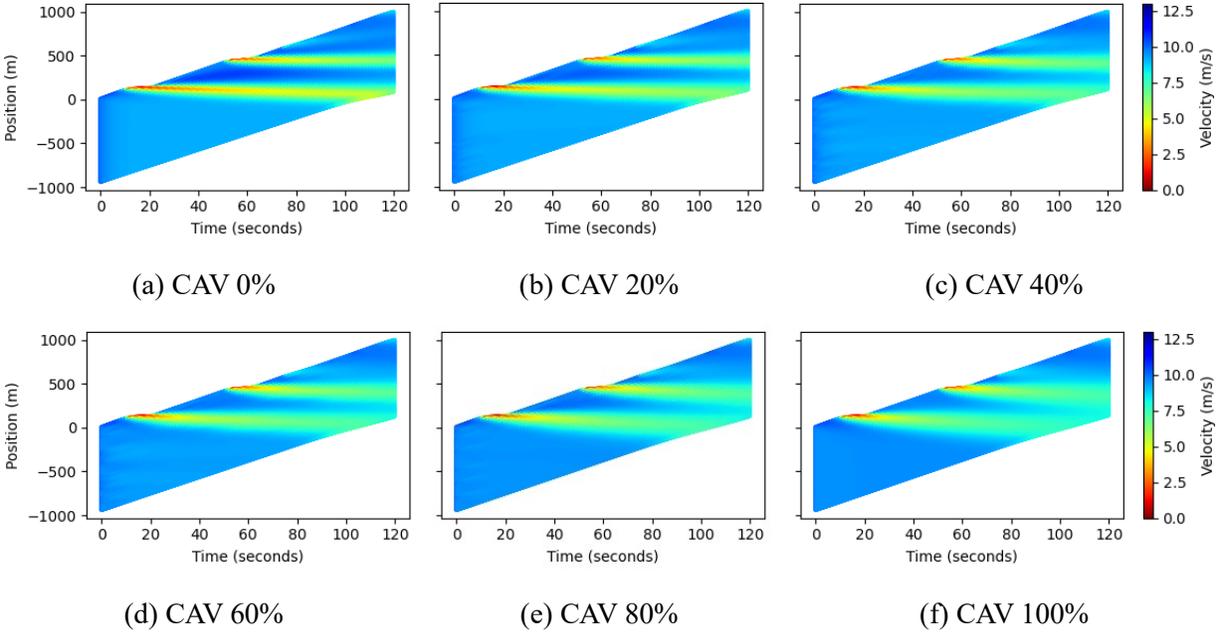

(a) CAV 0%  (b) CAV 20%  (c) CAV 40%

(d) CAV 60%  (e) CAV 80%  (f) CAV 100%

Figure 11 Velocity heatmap of the mixed platoon with various penetration rates.

## 5 CONCLUSION

This research presents a control framework based on PERPL for decentralized platoon control. It harnesses the interpretability and robustness of linear control policies alongside the flexible, multi-objective learning capabilities of reinforcement control policies. As such, this model exhibits high control precision, achieving stable headway and reduced traffic oscillation, and it demonstrates strong generalization capabilities, maintaining stability and safety in unseen domains. We applied our proposed model to decentralized control at the mixed traffic platoon level, utilizing CTG for cruising while considering actuator and communication delays. The model's performance was validated through artificially generated extreme scenarios and real-world trajectories. Experimental findings indicate that, both under artificial extreme conditions and with actual vehicle trajectories, our approach yields smaller headway errors and superior oscillation control compared to traditional linear and standalone RL methods. On a macroscopic traffic scale, traffic oscillations diminish as more CAVs adopt the PERPL framework, enhancing overall traffic dynamics.

For future research, we plan to leverage the safety features of the proposed method by testing it on both small-scale laboratory vehicles and larger vehicles. This approach will allow for a more realistic consideration of the cumulative effects of errors in perception, communication, and control systems that are commonly encountered in practical deployments. Moreover, we aim to enhance the capabilities of our framework by integrating it with advanced predictive models. This integration seeks to establish an end-to-end control system that reacts to immediate environmental inputs and anticipates future states. By combining real-time control adjustments with foresighted planning, the system could dynamically adapt to changes in traffic conditions, road layouts, and vehicle behaviors, significantly boosting its effectiveness in complex traffic scenarios.